\documentclass[
]{ceurart}

\usepackage{listings}
\usepackage{url}
\usepackage[capitalise]{cleveref}
\crefformat{footnote}{#2\footnotemark[#1]#3}
\usepackage{booktabs}
\begin{document}

\copyrightyear{2023}
\copyrightclause{Copyright for this paper by its authors.
  Use permitted under Creative Commons License Attribution 4.0
  International (CC BY 4.0).}

\conference{Scholarly QALD at ISWC 2023}

\title{NLQxform: A Language Model-based Question to SPARQL Transformer}

\author[1,2]{Ruijie Wang}[%
orcid=0000-0002-0581-6709,
email=ruijie@ifi.uzh.ch
]
\cormark[1]
\address[1]{Department of Informatics, University of Zurich, Switzerland}
\address[2]{University Research Priority Program ``Dynamics of Healthy Aging'', University of Zurich, Switzerland}

\author[1]{Zhiruo Zhang}[orcid=0009-0008-7115-9429,email=zhiruo.zhang@uzh.ch]

\author[1]{Luca Rossetto}[orcid=0000-0002-5389-9465,email=rossetto@ifi.uzh.ch]

\author[1]{Florian Ruosch}[%
orcid=0000-0002-0257-3318,
email=ruosch@ifi.uzh.ch
]

\author[1]{Abraham Bernstein}[%
orcid=0000-0002-0128-4602,
email=bernstein@ifi.uzh.ch
]

\cortext[1]{Corresponding author.}

\begin{abstract}
    In recent years, scholarly data has grown dramatically in terms of both scale and complexity.
    It becomes increasingly challenging to retrieve information from scholarly knowledge graphs that include large-scale heterogeneous relationships, such as authorship, affiliation, and citation, between various types of entities, e.g., scholars, papers, and organizations.
    As part of the Scholarly QALD Challenge, this paper presents a question-answering (QA) system called NLQxform, which provides an easy-to-use natural language interface to facilitate accessing scholarly knowledge graphs.
    NLQxform allows users to express their complex query intentions in natural language questions.
    A transformer-based language model, i.e., BART, is employed to translate questions into standard SPARQL queries, which can be evaluated to retrieve the required information.
    According to the public leaderboard of the Scholarly QALD Challenge at ISWC 2023 (Task 1: DBLP-QUAD — Knowledge Graph Question Answering over DBLP), NLQxform achieved an F1 score of 0.85 and ranked first on the QA task, demonstrating the competitiveness of the system.
    
\end{abstract}


\maketitle

\section{Introduction}
\label{sec:Introduction}

In the past decade, tremendous efforts~\cite{DBLP:journals/qss/WangSHWDK20,DBLP:journals/corr/abs-2205-01833,DBLP:conf/wims/AuerKPKSV18,banerjee2023dblp} have been devoted to organizing scholarly data in the form of knowledge graphs (KGs), which store heterogeneous relational information in an integrated and extensible manner, supporting querying with complex semantics.
The DBLP scholarly KG\footnote{\url{https://blog.dblp.org/2022/03/02/dblp-in-rdf/}} is one representative of such efforts.
It stores the entire DBLP\footnote{\label{footnote:dblp}\url{https://dblp.org/}} data in the format of RDF~\cite{DBLP:series/ihis/Pan09} and provides an endpoint\footnote{\url{https://dblp-kg.ltdemos.informatik.uni-hamburg.de/sparql}} for accessing the data with SPARQL~\cite{world2013sparql} queries.
Conventional tools for searching scholarly information, e.g., Google Scholar\footnote{\url{https://scholar.google.com/}}, DBLP bibliography\cref{footnote:dblp} and publisher-provided services\footnote{Examples include IEEE Xplore (\url{https://ieeexplore.ieee.org/Xplore/home.jsp}) and the ACM Digital Library (\url{https://dl.acm.org/}).}, only support text search with limited metadata-based filtering and sorting.
In contrast, scholarly KGs are more powerful and versatile, as they support queries with complex constraints (triple patterns in SPARQL) and operations (functions and modifiers in SPARQL). 
Nevertheless, scholarly KGs are significantly less commonly used than conventional tools.
A main hurdle is the complexity of the SPARQL language (cf. \cite{KAUFMANN2010377}).

To address the above issue, we developed a QA system that allows users to query the DBLP scholarly KG with natural language questions.
The system is called \textbf{NLQxform}, as it answers Natural Language Questions (\textbf{NLQ}s) using a \textbf{transform}er-based language model (LM). Specifically, NLQxform uses  BART~\cite{lewis2020bart}---a sequence-to-sequence model, which has achieved promising performance in natural language translation.
We first fine-tuned it with NLQ-SPARQL query pairs.
Given a question, NLQxform employs BART to generate a logical form that has the basic structure of the target SPARQL query.
Then, NLQxform links entities mentioned in the given question to their corresponding URLs in the underlying KG and corrects minor syntax and grammar errors in the logical form based on a SPARQL template base.
Finally, NLQxform generates explicit candidate SPARQL queries and evaluates them via the DBLP endpoint to retrieve answers.

This work is part of the Scholarly QALD Challenge,\footnote{\url{https://kgqa.github.io/scholarly-QALD-challenge/2023/}} which provides valuable benchmarking for QA over scholarly KGs.
We participated in both the Question Answering and Entity Linking sub-tasks in Task 1: DBLP-QUAD — Knowledge Graph Question Answering over DBLP.\footnote{\url{https://codalab.lisn.upsaclay.fr/competitions/14264}}
The remainder of this paper is structured as follows:
Related work on the two sub-tasks is introduced in \cref{sec:relatedwork}.
We elaborate on the NLQxform system in \cref{sec:system}.
The experimental setup and final results of NLQxform in the challenge are reported in \cref{sec:experiment}.
Finally, we conclude in \cref{sec:conclusion}.

\section{Related Work}
\label{sec:relatedwork}

In this section, we first survey QA approaches with a focus on previous QALD challenges~\cite{DBLP:conf/clef/CabrioCLNUW13,DBLP:conf/clef/UngerFLNCCW14,DBLP:conf/clef/UngerFLNCCW15,DBLP:conf/esws/UngerNC16,DBLP:conf/esws/UsbeckNHKRN17,UsbeckNCRN18,UsbeckGN018}.
Then, we present an overview of QA-related entity linking.

\textbf{Question Answering.}
\citet{banerjee2022modern} propose to employ pre-trained language models, such as T5~\cite{2020t5} and BART~\cite{lewis2020bart}, as well as Pointer Generator Networks~\cite{see2017get} to construct SPARQL queries using pre-identified entities and relations in given questions.
The fine-tuned T5-base model achieved an F1-score of $0.87$ on the DBLP-QuAD dataset~\cite{DBLP:journals/corr/abs-2303-13351}, which is also used for training and validation in this iteration of the Scholarly QALD Challenge.
WDAqua~\cite{diefenbach2017wdaqua,diefenbach2018wdaqua} participated in several of the past QALD challenges.
Diefenbach et al.~\cite{diefenbach2017wdaqua} propose a new approach by not focusing on the syntax but on the semantics of tokens in given questions to generate SPARQL queries.
First, recognized tokens are expanded to possible entities in the underlying KG.
Next, all possible SPARQL queries are constructed and ranked based on a linear combination of features.
Finally, top-ranked queries are selected and used to retrieve answers.
The authors integrated their approach to Qanary~\cite{both2016qanary}, modular architecture for QA systems, and evaluated it on QALD iterations three through seven~\cite{DBLP:conf/www/DiefenbachSM18}.
\citet{zou2014natural} recast the problem of NLQ to SPARQL translation with a graph-based solution.
They form subgraphs by mapping phrases from given questions to entities in the KG.
This relegates the problem of entity disambiguation to the query generation phase of their approach.
Finally, the previously created subgraphs are matched to the KG with heuristic rules to retrieve top SPARQL queries and answers.

\textbf{Entity Linking.}
DBLPLink~\cite{DBLP:journals/corr/abs-2309-07545} is a web application specifically developed for entity linking over the DBLP scholarly KG.
It employs pre-trained language models to identify entity mentions in given questions and links them to KG entities using labels and entity embeddings.
\citet{steinmetz2023entity} presents an approach where input phrases are parsed to an abstract meaning representation graph based on a pre-trained BART model.
From this representation graph, named entities are extracted and subsequently mapped to the KG using various information, including alternative labels, links, and node degrees.
The system was evaluated on QALD-9~\cite{DBLP:conf/semweb/UsbeckGN018}, outperforming competing systems.
\citet{diomedi2022entity} propose a novel system called ElNeuKGQA, which combines entity linking with neural machine translation.
To alleviate the issue of unseen words, the authors utilize a neural machine translation model to generate template queries wherein out-of-vocabulary words are replaced by placeholders.
These are then filled by identified entities afterward in an “entity filling” phase, where sequence labeling is used to determine the role of placeholders.
EARL~\cite{dubey2018earl} unifies entity linking and relation linking.
To this endeavor, the authors present two different approaches.
The first approach recasts the linking problem as the Generalised Traveling Salesman Problem, while the second uses an xgboost~\cite{chen2016xgboost} classifier based on the connection density as determined by links in the KG.
The two approaches also differ in that the former does not require training data, while the latter does.
EARL was evaluated on the QALD-7 dataset~\cite{DBLP:conf/esws/UsbeckNHKRN17} and achieved superior performance over its previous competitors.

\section{The NLQxform System}
\label{sec:system}

\begin{figure}[t]
    \centering
    \includegraphics[width=0.9\textwidth]{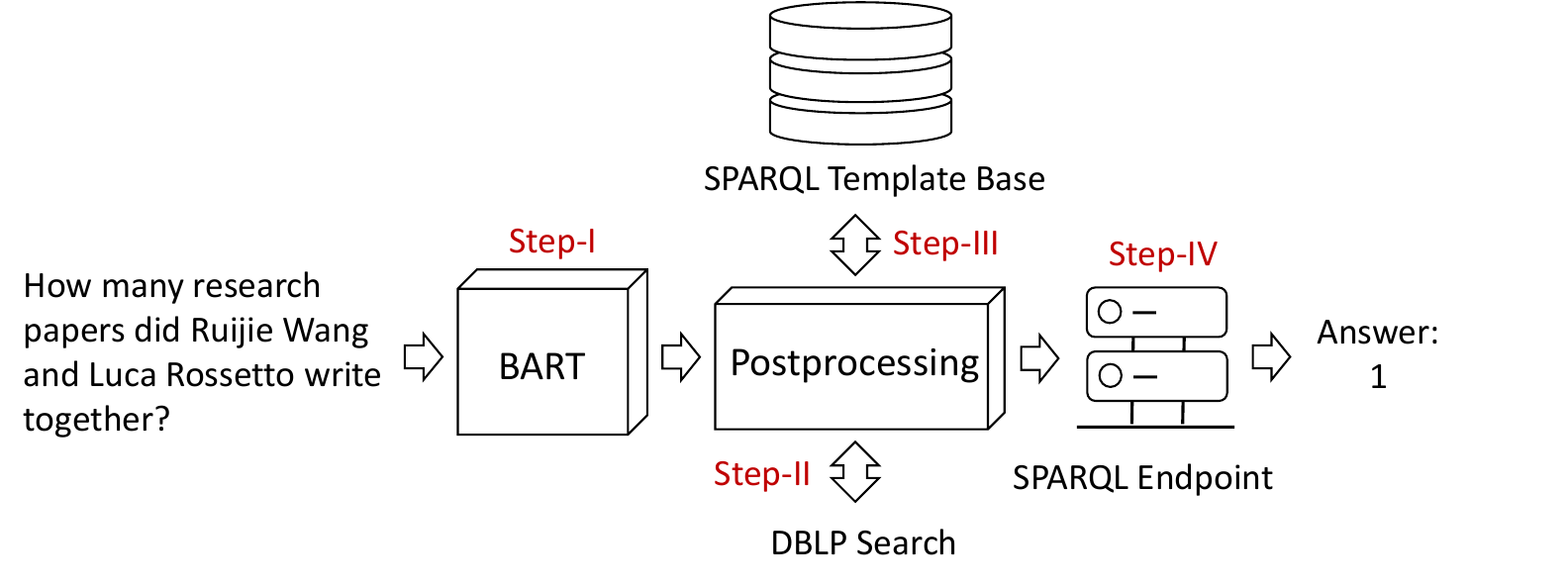}
    \caption{An overview of the NLQxform system and its four steps question-answering process.} 
    \label{fig:system}
\end{figure}

In this section, we elaborate on the NLQxform system.
An overview of the system structure is presented in \cref{fig:system}, where a natural language question is answered via four steps:

\paragraph{Step-I: BART-based question to logical form translation.} 
In the first step, a BART ~\cite{lewis2020bart} model (facebook/bart-base\footnote{\url{https://huggingface.co/facebook/bart-base}}) is used to translate the given question into a logical form that is very close to the final standard SPARQL query, except that entities in the logical form are still natural language mentions instead of URLs of entities.
For example, given the question ``\textit{how many research papers did Ruijie Wang and Luca Rossetto write together},'' the BART model is expected to generate the logical form: \begin{verbatim}
SELECT COUNT(DISTINCT ?answer) AS ?count 
WHERE {
   ?answer <https://dblp.org/rdf/schema#authoredBy> <Ruijie Wang> .
   ?answer <https://dblp.org/rdf/schema#authoredBy> <Luca Rossetto> .
}
\end{verbatim}
Please note that the adopted BART model was not pre-trained on the question-to-SPARQL translation task.
Therefore, it needs to be fine-tuned with training question-SPARQL query pairs.\footnote{Please refer to \cref{sec:experiment} for information about training questions.}
Also, we make the following modifications to better suit the task:
First, basic elements of SPARQL syntax, including clauses, functions, and modifiers (e.g., ``SELECT'', ``COUNT'', and ``ORDER BY''), as well as some tokens frequently used in SPARQL queries (e.g., ``\{'', ``\}'', ``('', ``)'', and ``.'') are added as special tokens to the tokenizer of the language model (LM).
Second, considering that the number of relations in the underlying knowledge graph is limited, we also add all relations as special tokens to the LM.\footnote{Only 12 relations are used in our system for this challenge.} 
Therefore, the LM is able to recognize relations in given questions and directly generate logical forms with relation URLs.

\paragraph{Step-II: DBLP search-based entity linking.}
In the above step, entity mentions in the given question are recognized and used to construct the logical form.
In this step, we search specific URLs of recognized entities using DBLP search APIs.\footnote{\url{https://dblp.org/faq/How+to+use+the+dblp+search+API.html}}
For example, regarding entities \texttt{<Ruijie Wang>} and \texttt{<Luca Rossetto>}, their correct URLs \texttt{<https://dblp.org/pid/57/5759-3>} and \texttt{<https://dblp.org/pid/156/1623>} can be obtained via requesting \texttt{\url{https://dblp.org/search/author/api?q=Ruijie\%20Wang}} and \texttt{\url{https://dblp.org/search/author/api?q=Luca\%20Rossetto}}.

\paragraph{Step-III: SPAQRL template-based query correction.}
This step is proposed to alleviate possible syntax and grammar errors in the generated logical forms.
During training, we utilize ground-truth SPARQL queries to construct a SPARQL template base.
For example, assuming the above question is a training question, the following template can be added:
\begin{verbatim}
SELECT COUNT(DISTINCT ?answer) AS ?count 
WHERE {
   ?answer <https://dblp.org/rdf/schema#authoredBy> <entity_1> .
   ?answer <https://dblp.org/rdf/schema#authoredBy> <entity_2> .
}
\end{verbatim} 
It provides a gold example for future questions that require the count of publications authored by two given scholars. 
During inference, regarding each generated logical form, we find top-\textit{3} similar templates from the template base according to string similarity and pass them to the next step.

\paragraph{Step-IV: SPARQL endpoint-based answer retrieval.} 
In this step, we fill in retrieved query templates with entity linking results to generate specific SPARQL queries and evaluate them via the SPARQL endpoint\footnote{\url{https://dblp-kg.ltdemos.informatik.uni-hamburg.de/sparql}} of DBLP scholarly KG to retrieve answers.
For example, the generated SPARQL query for the above question is expected to be
\begin{verbatim}
SELECT COUNT(DISTINCT ?answer) AS ?count 
WHERE {
   ?answer <https://dblp.org/rdf/schema#authoredBy>
                                     <https://dblp.org/pid/57/5759-3> .
   ?answer <https://dblp.org/rdf/schema#authoredBy>
                                     <https://dblp.org/pid/156/1623> .
}
\end{verbatim}
The final answer retrieved with this query is \textit{1}.\footnote{This result is with respect to the snapshot of the KG used by the endpoint.}
It is worth mentioning that there could be several candidate URLs returned for one entity from Step-II.
In this case, we follow the order of these URLs returned by the DBLP API and adopt the first generated SPARQL query that can return an answer in the endpoint.

\section{Experiments}
\label{sec:experiment}

In this section, we report the experimental setup and evaluation results of NLQxform in the Scholarly QALD Challenge (Task 1: DBLP-QUAD — Knowledge Graph Question Answering over DBLP).\footnote{\url{https://codalab.lisn.upsaclay.fr/competitions/14264}}

\textbf{Dataset.} 
The employed QA dataset is DBLP-QuAD~\cite{DBLP:journals/corr/abs-2303-13351}, which includes natural language questions posed over the DBLP scholarly KG.
The training and validation sets of DBLP-QuAD, respectively, include 7,000 and 1,000 questions with annotations of correct answers, ground-truth SPARQL queries, and entity-linking results.
They were used for the fine-tuning of the BART model.
In the final phase of the challenge, 500 newly generated test questions\footnote{\url{https://github.com/debayan/scholarly-QALD-challenge/blob/main/2023/datasets/codalab/finalphase/dblp-kgqa/dblp.heldout.500.questionsonly.json}} without answer annotations were provided for system evaluation.

\textbf{Evaluation Setup.} 
The entity linking results from Step-II and final answers of NLQxform were submitted to the Codalab platform,\footnote{\url{https://codalab.lisn.upsaclay.fr/competitions/14264\#participate}} where final results regarding entity linking and QA were automatically calculated.
F1-score was used as the criterion for both tasks.

\begin{table}[]
\caption{The final evaluation results regarding entity linking and question answering. (best performance in \textbf{bold}, second best \underline{underlined})}
\label{tab:eval_results}
\begin{tabular}{@{}ccc@{}}
\toprule
Submission           & F1 Score (Entity Linking)  & F1 Score (Question Answering)  \\ \midrule
ID-544291            & 0.8283 & 0.0000 \\
ID-544863            & \underline{0.8320} & 0.0000 \\
ID-557116            & \textbf{0.8353} & 0.0000 \\
ID-545920            & 0.7100 & 0.0018 \\
ID-556670            & 0.6235 & 0.2175 \\
ID-547129            & 0.0000 & \underline{0.6619} \\
ID-557036 (NLQxform) & 0.7961 & \textbf{0.8488} \\ \bottomrule
\end{tabular}
\end{table}

\textbf{Final Results.}
We report the final results of public submissions\footnote{\url{https://codalab.lisn.upsaclay.fr/competitions/public_submissions/14264}} in \cref{tab:eval_results}.
The following can be observed:
\begin{itemize}
    \item 
    On the entity linking task, NLQxform is ranked fourth regarding the F1 score. 
    Compared to the best-performing system, the F1 score of our system is 4.7\% lower.
    The main reason is that there is a lack of explicit supervision for the recognition of entities in Step-I, as the fine-tuning objective is the matching of whole queries without emphasis on entities.
    Also, due to an implementation issue, the linked entities that cannot result in valid SPARQL queries in Step-IV were not removed from our submission.
    Nevertheless, this shows that there is still room for improvement in our system.
    In addition, the precision and recall regarding entity linking are 0.81 and 0.79, respectively.

    \item 
    On the QA task, NLQxform achieved the best performance, significantly improving over the second-best system (+28.2\%).
    This demonstrates the overall effectiveness of our system and shows that the LM was successfully fine-tuned to learn the syntax and grammar of SPARQL. 
    In addition to the above results, the precision and recall of our system on QA are 0.83 and 0.87, respectively.

    \item 
    The QA task seems more challenging than the entity linking task in this challenge.
    On entity linking, most submissions achieved an F1 score above 0.6.
    However, on QA, only two systems are above 0.6, and the other two systems with a result are both below 0.3.
    This further demonstrates the competitiveness of NLQxform.
    
\end{itemize}

\section{Conclusion and Outlook}
\label{sec:conclusion}

In this paper, we present our QA system, NLQxform, which fine-tunes a transformer-based BART model and answers NLQs over the DBLP Scholarly KG via four steps: 1. BART-based question to logical form translation, 2. DBLP search-based entity linking, 3. SPARQL template-based query correction, and 4. SPARQL endpoint-based answer retrieval.
NLQxform participated in the Scholarly QALD Challenge and ranked first on the QA task over DBLP-QUAD with a significant improvement over the second-best system (+28.2\%), demonstrating the system's effectiveness and competitiveness.

A limitation of NLQxform is that the recognition of entities is not explicitly emphasized during the fine-tuning of the BART model, which led to an inferior entity linking performance.
We will investigate this issue in future work.
Also, some technical details of the system are not fully presented in this paper due to the limitation of space and readability considerations.
They will be more comprehensively explained and analyzed in an extended version of this work.

\begin{acknowledgments}
    This work was partially funded by the Digital Society Initiative of the University of Zurich, the University Research Priority Program ``Dynamics of Healthy Aging'' at the University of Zurich, 
    and the Swiss National Science Foundation through Projects \href{https://data.snf.ch/grants/grant/184994}{``CrowdAlytics''} (Grant Number 184994) 
    and \href{https://data.snf.ch/grants/grant/202125}{``MediaGraph''} (Grant Number 202125). 
\end{acknowledgments}

\bibliography{refs}


\end{document}